\documentclass[11pt]{article}
\usepackage{acl2017}
\usepackage{times}
\usepackage{url}
\usepackage{latexsym}
\usepackage{xspace}
\usepackage{booktabs}
\usepackage{color}
\usepackage{graphicx}
\usepackage{tabulary}
\usepackage{enumitem}
\usepackage{url}
\usepackage{amsfonts}

  {\begin{itemize}[topsep=0pt, partopsep=0pt] %
    \setlength{\itemsep}{0pt}%
    \setlength{\parskip}{0pt}%
    }%
  {\end{itemize}}
  \usepackage{booktabs}
\usepackage{subcaption}
\usepackage{lipsum}

  \usepackage{color}
  {\begin{enumerate}â \footnotesize%
    \setlength{\itemsep}{0pt}%
    \setlength{\parskip}{0pt}%
    }%
  {\end{enumerate}}


\usepackage{microtype}
\usepackage{multirow}
\usepackage{verbatim}
\usepackage{amsmath,amsthm,amssymb}
\usepackage{array}
\usepackage[scaled=0.86]{helvet}
\usepackage{ifthen}
\usepackage{courier}
\usepackage[boxed]{algorithm}
\usepackage{caption}
\usepackage{algpseudocode}
\aclfinalcopy

\belowdisplayskip \abovedisplayskip

\usepackage{amsmath}
\newcommand{\sts}{{{\textsc{Seq2Seq}}}\xspace}

\newcommand{\todo}[1]{\textcolor{red}{#1}}

\title{Data Distillation for Controlling Specificity in Dialogue Generation}
\author{
{\bf } 
Jiwei Li{\normalfont ,}\; Will Monroe \and Dan Jurafsky 
\\ 
Computer Science Department\\
 Stanford University, Stanford, CA, USA,\\
{\tt jiweil,wmonroe4,jurafsky@stanford.edu} 
}

\begin{document}
\maketitle
\begin{abstract}

People speak at different levels of specificity in different situations.\footnote{Depending on their knowledge, interlocutors, mood, etc.} A conversational agent should have  this ability  
 and  know when to be specific and when to be general.

We propose an approach that gives a  neural network--based
conversational agent this ability. Our approach involves
alternating between   \emph{data distillation}
and model training
: removing
training examples that are closest to the responses most commonly produced
by the model trained from the last round and then retrain the model on the remaining dataset.
Dialogue generation models trained with different degrees of data distillation
manifest different levels of specificity.
 
We then train a reinforcement 
learning system for selecting among this pool of generation models,
to choose the best level of specificity for a given input.
 Compared to the original generative model trained without distillation, the proposed 
 system is capable of generating
 more interesting and higher-quality responses, in addition to
 appropriately adjusting specificity depending on the context.

Our research constitutes a specific case of a broader approach involving
training multiple subsystems from a single dataset distinguished by differences in a specific property one wishes to model. We show that from such a set of subsystems,
one can use reinforcement learning to build a system that tailors its output to different input contexts at test time. 

\end{abstract}

\section{Introduction}
\label{sec:intro}
People use different levels of specificity in their language depending on many
factors about the context of a conversation: one's interlocutor, one's mood,
how familiar one is with the topic discussed,
how well one understands the other's utterances, and so forth all influence the
decision to respond with generics or specifics.
A good dialogue agent should have a similar ability to vary the level of
specificity of the responses it generates in an input-dependent way. 

When humans speak, 
we can imagine that each has a series of language  models  in his mind, 
each of which is able to generate a sensible response, but
which differ in  specificity. One picks the appropriate model according to the current situation (whether one understands the input utterance, whether one is interested in the topic, etc.)
and generates a dialogue utterance using the selected model. 
Motivated by this line of thinking, we ask whether a conversational agent could consider a pool of dialogue models that vary in language specificity and pick the best one for producing a response to any given input.

One seemingly straightforward approach would be to split the training data by language specificity and train  separate generation models on each split. However, this requires
classifying data by
text specificity, a problem which poses significant challenges. Language specificity has been historically studied 
for noun phrases, and a few specificity-indicative features have been identified, such as singular terms, negations, or actual/non-actual moods \cite{encc1991semantics,lyons1995linguistic}. 
However, there is no generally agreed criterion for defining the level of specificity of an arbitrary unit of natural language, let alone automatically generating sequences to have different levels of specificity.

In this paper, 
we propose an iterative \emph{data distillation} approach for addressing this issue.\footnote{The  model  is inspired by the concept of distillation in chemistry, 
which
separates chemical mixtures by
gradually increasing the temperature to a point
 at which one or more compounds in the mixture will vaporize.}
The proposed system operates as follows: a neural sequence-to-sequence generation (\sts) model is first trained and used to generate (decode) responses to inputs in a dataset. 
A list of the most common responses
 is constructed, and training examples with outputs that are semantically close to these common responses are removed (distilled). 
This  
process is then repeated 
 by training another \sts model (from scratch) on the remaining data, decoding using the trained model, collecting generic responses and distilling more data. 
As the process iterates, responses that 
are generic
 are gradually distilled, 
and the trained models gradually increase in specificity. 

At the end of the entire data distillation process, we are presented with a pool of generation models, all of which are able to produce sensible responses to input messages but differ in degree of specificity. 
This pool of  models is analogous to specificity-varying models in a human's mind. When presented with an input dialogue message, the dialogue system needs to pick one model out of the pool, 
Which model to choose depends on how well the bot understands the input message, how knowledgeable it is regarding the topic discussed, etc.\footnote{We leave handling other factors that should influence specificity (such as the current mood of the bot and non-linguistic characteristics of the interlocutor) for future work. }
To imbue the agent with this ability, we use reinforcement learning to train a model to pick the an appropriate level of specificity by selecting one of pre-trained generative models from the pool.

Experimental results show that  models trained from different rounds of data distillation exhibit a clear spectrum of specificity. 
Models trained in early rounds of data distillation yield better responses. 
We also show that
the reinforcement learning model is able to choose levels of specificity that
are appropriate for a variety of inputs.

Our research constitutes a specific case of a broader approach involving
training multiple subsystems from a single dataset distinguished by differences in a specific property one wishes to model (here, specificity), 
especially when this property is hard to model in a supervised learning setting. We show that from such a set of subsystems,
one can use reinforcement learning to build a system that tailors its output to different input contexts at test time. 
\section{Related Work}
\paragraph{Generic responses in open-domain dialogue}
End-to-end dialogue systems 
\cite{ritter2011data,serban2016,vinyals2015neural,serban2016hierarchical,serban2016multi,RE,mei2016coherent}, tend to generate highly generic and commonplace responses \cite{sordoni2015neural,mou2016sequence}.
The goal of controlling output specificity is closely related to recent attempts  to address this issue.
 \newcite{li2015diversity} propose using mutual information as
an alternative
 training objective function in place of maximum likelihood, in which an N-best list generated by $p(t|s)$ is reranked by the backward probability $p(s|t)$. 

The aim of this work is more general:  
instead of attempting to always avoid generic responses, our goal is to provide the system with the flexibility to generate responses at different levels of specificity. 
Blindly avoiding generating generic responses does not reflect how humans speak: we do say dull, generic things like {\it I don't know what you are talking about}, to communicate that we indeed do not understand part of the conversation, or to dismiss something as incorrect or nonsensical. A good dialogue system should have the ability to decide when to say generic things and when not to. 

\paragraph{Data manipulation}
The
idea of training with data distillation  is inspired by a variety of work 
in
the active learning
and subdata selection literature,  
the key idea of which is to 
select a subset of a large
dataset to train a classifier with minimal
performance loss,
for when the training dataset is extremely large or training is extremely time-intensive
\cite{wei2015submodularity,zheng2014submodular,prasad2014submodular,ghahramani2013scaling,iyer2013submodular}. The proposed system differs from these subdata selection methods in both goals and implementation: we combine a series of models trained on different subsets of data,
with the goal of increasing model performance rather than preserving the model's performance while reducing the size of the training data.

The system we propose is also related to 
data manipulation
strategies
 such as boosting \cite{breiman1996bias}, a type of ensemble method \cite{dietterich2002ensemble,zhou2002ensembling,krogh1995neural}
that 
uses subsets of the original data to produce a series of  models and then "boosts" their performance by combining them together; and bagging \cite{breiman1996bagging}, which
generates additional data for training using the original dataset  to produce multisets of the same size as the original data, decreasing training variance.

 \section{Data Distillation}
In the section, we describe the proposed data distillation model in detail. 
We use OpenSubtitles \cite{tiedemann2009news} as our training dataset.\footnote{OpenSubtitles is a large,
noisy, open-domain dataset of lines from movie scripts.
The noise in the dataset is largely due to the lack of speaker labels for
lines of the subtitles. Following Vinyals et al. (2015), we
train our models to predict the current line given the preceding ones,
assuming that each line constitutes a full speaker turn and
that consecutive turns belong to the same conversation. Both assumptions
are occasionally untrue but yield reasonable results.}

\subsection{Distilling common responses}

We first use the following simple example to illustrate the core idea of our system: 
consider a model that predicts a multinomial distribution over an output variable (e.g., which fruit to choose).
The probability of picking 
 {\it apple} is 0.3, {\it orange} 0.25, {\it blueberry} 0.15, {\it blackberry} 0.15, and {\it raspberry} 0.15.
 Outputs that are generic are usually highly probable, since the high diversity of specific outputs results in each having smaller probability mass.
 We thus treat {\it apple} as the most generic fruit, and the various berries as more specific. 
Maximum likelihood estimation at test time will lead  the model to always 
choose {\it apple}, since it has the largest probability. 
Observing that {\it apple} is the most common output, we will remove all {\it apple}s from the training set
and retrain the model, which will pick {\it orange} this time, since it has the greatest probability after {\it apple}s are removed. We then remove {\it orange}s and repeat the process. 
With successive iterations of this distillation process, we will gradually obtain models that produce more specific outputs. 

In the context of dialogue response generation, our approach works as follows: for each iteration, we first train a \sts model using attention \cite{bahdanau2014neural,luong2015effective}   on the original training set.
Next, we use the trained model to decode responses to a number of input examples. We decode only a subset of the training set, 1 million responses in total.
One could also use a held-out dataset for decoding, but  the source of input messages is fairly unimportant in identifying the most frequent responses. We use greedy decoding (beam search with beam size 1).
We then 
collect the most common responses in a list, denoted by $L$.
A response is considered generic if its frequency of occurrence exceeds a threshold, which is empirically set to 100 in this work. 
We then compare each response in the training data to each highly frequent 
response from the list $L$
and assign
a relevance score $R(e)$ to each training example $e$ based on
the cosine similarity between $e$ and the sequence most similar to it in $L$:
\begin{equation}
R(e)=\text{max}_{e'\in L}~~ \text{cos}(e,e')
\label{relevance}
\end{equation}
We use the encoder part of the trained encoder-decoder model to map these sequences to vector representations, which are used to compute the cosine similarity. In this way, sentences that are semantically similar to frequent responses are assigned high relevance scores.\footnote{Other options include skip-thought vectors \cite{kiros2015skip} and bag-of-word representations. We find using the trained encoder works decently well.}
We then remove (distill) examples from the training data with the highest relevance scores\footnote{The amount to remove is empirically set to 8--10\%.}
and retrain a new \sts model on the data that remains. An outline of the distillation algorithm is shown in Algorithm~\ref{distill}.

\algrenewcommand{\algorithmicrequire}{\textbf{Input:}}
\algrenewcommand{\algorithmicensure}{\textbf{Output:}}
\newcommand{\To}{{\bf to }}

\begin{algorithm}[t]
\small
\begin{algorithmic}
\Require training data $D$
\Ensure sequence of trained models $M$
\State
\State $M \gets \emptyset$
\For {$i \gets 1$ \To $N = 8$}
  \State train a \sts model $m$ on $D$ until convergence
  \State $M \gets M + m$
  \State decode subset of input messages in $D$ using model $m$
  \State collect top frequent decoded responses $L$
  \ForAll {instances $e \in D$}
    \State compute relevance score $R(e)$ using Eq. \ref{relevance}
  \EndFor
  \State $D^\neg \gets {}$top examples by $R(e)$
  \State distill $D^\neg$:  $D \gets D-D^\neg$
\EndFor
\State \Return $M$
\end{algorithmic}
\caption{A brief summary of the proposed data distillation algorithm.}
\label{distill}
\end{algorithm}

\subsection{Choosing a specificity model}
The data distillation process produces a pool of \sts models, 
each trained on the dataset remaining after a different data distillation round.
When presented with an input message at test time, the system has to decide which generation model from the pool to use to decode a response to the input. 
We repeat the data distillation process 8 times, which means we have 8 models in the pool to choose from.\footnote{It requires two Tesla K40 GPUs to fit the 8 models in memory.}
The system should have the ability to choose different models in response to properties of different inputs. For example, a good dialogue system should give concrete responses when asked things that it is sure about, but generic ones when the input message is difficult to understand. 

We use reinforcement learning to train a model to make this choice.
Given an input message $X$ from a held-out dataset, 
we parameterize the action of choosing the generative model with index $i$ from the pool $G=\{g_i\}$ of \sts models trained with data distillation
 as a policy network $\pi(g_i|X)$, which produces a distribution over $|G|$ classes. 
 To compute the distribution,
we first map the input $X$ to a vector representation $h_X$ using an LSTM and then map $h_X$ to a policy distribution over different $g_i\in G$ using a softmax function:
\begin{equation}
\pi(g=g_i|X)=\frac{\exp( h_X^T\cdot h_{g_i})}{\sum_{j=1}^{j=|G|}\exp( h_X^T\cdot h_{g_j})}
\label{softmax}
\end{equation}
 where $h_{g_i}$ is an output vector for each model $g_i$ that is randomly initialized and then trained. 
Given an action, namely a choice of a generative model $g_i$, we start decoding given the input message $X$ using that model. 
Decoding generates an output response $y$, which yields a reward $R(y)$ evaluating response quality according to some metric. 
The reward signal $R(y)$ is used to train the policy network.

We use the REINFORCE algorithm \cite{williams1992simple}, a kind of policy gradient method, to find the optimal policy by maximizing the expected reward $E_{\pi(g_i|X)} [R(y)]$. The expectation is approximated by sampling from $\pi$ and the gradient is computed using the likelihood ratio \cite{aleksand1968stochastic}:
 \begin{equation}
\begin{aligned}
\nabla E(\theta)\approx [R(y)-b] \nabla\log \pi(g_i|X))
\end{aligned}
\label{hahaha}
\end{equation}
where $b$ denotes a baseline value. \footnote{The baseline value is estimated using another neural model.
We refer the readers to \newcite{ranzato2015sequence} and \newcite{zaremba2015reinforcement} for more details.}

\paragraph{Adversarial evaluation for reward calculation}
One remaining question is how to assign a reward $R$ to a generated response $y$ given the input $X$, which boils down to the fundamental question of how to evaluate the general quality of a generated response. 
 Dialogue quality is traditionally evaluated \cite[e.g.]{sordoni2015neural} using word-overlap metrics such as BLEU and METEOR scores used for machine translation, which have recently been found to correlate poorly with human evaluations \cite{liu2016not}. 
 Recent work has begun using more flexible and reliable evaluation metrics;
 automatic prediction of human ratings \cite{hey} is one such metric, but
 this approach requires a large amount of human labeling effort to train a
 prediction model. 

We employ adversarial evaluation \cite{add,kannan} for reward calculation.
The idea of adversarial evaluation, first proposed by 
\newcite{bowman2015generating}, is to 
train a discriminator (or evaluator)
function
to labels dialogues as machine-generated (negative) or  human-generated (positive),
a binary classification task. 
For our system, we use positive examples taken directly from training dialogues, while negative examples are decoded using generative models from different rounds of data distillation.
To be specific,
for each input message, we randomly sample a \sts model from the pool to decode a response to the input and use the response as a negative example. The evaluator is a hierarchical neural model \cite{serban2015hierarchical}:   dialogue utterances (i.e., source messages and responses) are first mapped to vector representations using an LSTM. Another LSTM is applied to the sequence of utterance representations to produce a dialogue representation, which is then fed to a binary classifier. 

Given a pre-trained evaluator $D$, an input source $X$ and a machine generated target $y$
decoded by the chosen generative model,
 the reward $R$ used to update the policy $\pi$ is the probability that the evaluator $D$ assigns to labeling $y$ as a human-generated response. The policy update influences the choice of generative model for decoding the current input $X$. 
We refer readers to \cite{add} for more details about the adversarial evaluation. 
\subsection{Stochastic Greedy Sampling}
Language specificity also relates to language diversity. 
Utterances with lower levels of diversity are usually generic because generic responses are usually generic in the same way. 
Modeling diversity also provides an indirect way to handle the issue of specificity. 

Moreover,
there is a degree of randomness in human language production: in the real world, if we ask a person the same question twice, even with the same environment and surroundings, it is unlikely that the person will give the same answer both times. 
Sampling from the distribution not only better mimic the way humans generate tokens, but also provides a way to handle the issue of language specificity . 

One simple solution is to sample directly from the distribution $p(y|x)$ in all cases. However,
we observe that sampling leads to 
 incoherent, ungrammatical, or even irrelevant responses. 
We expect there to be a sweet spot on the spectrum of randomness, between full sampling on one end and greedy or beam search on the other.\footnote{Since greedy decoding has been shown to generate \ higher-quality responses than beam search in dialogue response generation \cite{li2015diversity}, we focus on greedy decoding. However, all algorithms  can be easily adapted to use beam-search decoding. 
}

We propose a straightforward algorithm called Stochastic Greedy Sampling, in which
instead of sampling from the full distribution over all candidate tokens, the model only samples from the few (e.g., 5) words with the highest probability. 
The model provides with both the flexibility of incorporating randomness and the rigidity of adhering to a pre-trained generation model at the same time. 

Again, we use Adversarial Evaluation for comparing purposes. 
We report {\it AdverSuc} and {\it machine-vs-random}  proposed by \newcite{kannan}.
{\it machine-vs-random} denotes the  the accuracy of distinguishing between machine-generated responses and randomly sampled responses using a machine evaluator, trained in a way similar to the evaluator in {\it AdverSuc}.
Table \ref{whereToSample} 
presents results
for {\it AdverSuc} and {\it machine-vs-random} results  for {\it greedy decoding}, {\it pure sampling} and the proposed 
{\it stochastic greedy model}. 
As can be seen, {\it sampling} all the time obtains the best score for {\it AdverSuc}, but also extremely low score for {\it machine-vs-random} accuracy, which indicates the inferiority of the always sampling strategy. 
The proposed {\it stochastic greedy} model perform better than
always taking greedy actions as in {\it greedy}. 
 This indicates that properly combining greedy search and sampling will potentially lead to better results. 
 
\begin{table}[!ht]
\small
\centering
\begin{tabular}{ccc}
Model&AdverSuc&machine-vs-random \\\hline
greedy&0.042&0.935\\
pure sampling&0.384&0.642\\
stochastic greedy&0.058&0.933\\
\end{tabular}
\caption{Adversarial evaluation results for different greedy vs.\ sampling decoding strategies.}
\label{whereToSample}
\end{table}

\section{Experimental Results} \label{experiments}
In this section, we present the results of experiments.
\subsection{Comparing generative models from different iterations}
It is interesting to first compare the generative models
and the remaining training data
 from each of the 8 rounds of data distillation. 
We use {\it Iter}+{\it N} to denote the generation model trained on the dataset after $N$ repetitions of data distillation. 

\begin{table}
\small
\centering
\begin{tabular}{cccccc}\\\hline
iter&data size&ppl& oracle-ppl&div-1&div-2  \\\hline
1 & 45.2M&33.2&33.2 & 0.65$\%$ & 1.57$\%$ \\
2&  40.6M&33.3&32.3 & 0.92$\%$& 2.81$\%$ \\
3&  35.7M&33.7&31.6 & 1.18$\%$& 3.22$\%$ \\
4&  32.7M&34.3&31.2 & 1.44$\%$& 3.60$\%$ \\
5&  30.1M&35.0&30.8 & 1.87$\%$& 3.94$\%$ \\
6&  27.9M&35.5&30.7 & 2.21$\%$ & 4.32$\%$ \\
7&  25.5M&36.7&30.5 & 2.72$\%$& 4.65$\%$ \\
8&  22.8M&37.2&30.3 & 3.10$\%$& 5.01$\%$ \\\hline
\end{tabular}
\caption{Training set size (examples) after data distillation in each iteration and perplexity (ppl) and $n$-gram diversity scores (dis-$n$) of the trained generative models on the development set. }
\label{ppl-diverse}
\end{table}

\begin{table*}[!ht]
\footnotesize
\centering
\begin{tabular}{p{1cm}p{6cm}p{1cm}p{6cm}}
Count&Response&Count&Response\\\hline
\multicolumn{2}{c}{{\bf Iter1}}&\multicolumn{2}{c}{{\bf Iter2}} \\\hline
145575&i don 't know what you are talking about .&54227&i 'm not in the mood .\\
84435&i 'm not going to let you go .&29559&i 'm sorry about the way i acted .\\
36032&i 'm sorry i didn 't mean to offend you .&22987&you 're not in the mood .\\
23890&i 'm not so sure .&21392&i 'm gonna take a look at the new york times .\\
19405&i don 't know what to say .&20380&i 'll be there in a minute .\\
16888&i 'm not going to let you go !&14736&i 'm gonna take a look at this .\\
16048&that 's a good idea .&13753&i 'll get the money .\\
12782&i don 't know what to do .&13013&i 'm gonna take a shower .\\
11840&i 'm not going to be able to do that .&11746&i 'm in the middle of a war .\\
11604&i 'm sorry i can 't help you .&10130&you 're not getting any sleep .\\
11254&i 'm sure you 're right .&9996&i 'm gonna take a look at the other side .\\
9474&you don 't know what you are saying .&9644&i 'm sorry about the way you did .\\
9471&i 'm not going to tell you .&9169&i 've been doing a lot of things .\\
8905&i 'm not sure i can do it .&7837&you 're a dead man .\\
7905&i have no idea .&5320&i was just getting a little tired of it .\\\hline
\multicolumn{2}{c}{{\bf Iter3}}&\multicolumn{2}{c}{{\bf Iter4}} \\\hline
41139&i 'm not an idiot .&30378&i 'm not from around here .\\
34738&i 'm not an expert on this .&26705&i 'm not from the future .\\
20252&i 'm sorry but i 'm not an expert on this .&9923&i was just talking to my wife .\\
16275&i 've got some bad news for you .&9012&i 'm not doing this .\\
16081&i 'll get you a new suit .&8573&you 're a goddamn liar .\\
13007&i 'm not an idiot !&7424&i 'll be on the way .\\
11254&i 'm gonna make a big deal out of this .&6919&i 'm sorry ma 'am .\\
6532&i 'm just an ordinary man .&5546&i 'm going back to the hotel .\\
5724&i 'm not an expert on the police .&4569&i 'll be on my way .\\
5604&i 'm not an expert on the subject .&4555&i 'm not staying here .\\
5168&i 'm not your enemy !&4416&you 're a goddamn genius .\\
4963&i 'm not an expert on the law .&4184&i 'm a little tired .\\
4454&i 'm gonna need some more help with this .&4183&i 'm gonna take a look at this .\\
4342&i was just about to get my hands on the wall .&4103&he 's a bit of a jerk .\\
3969&i can 't believe you 're still alive .&3819&he 's a bit of a pain in the ass .\\\hline
\end{tabular}
\caption{Most frequent responses generated using greedy search at the end of 1--4 rounds of data distillation. ``Count'' indicates the number of occurrences of a response in 1 million decoded outputs. }
\label{top-frequent}
\end{table*}
\begin{table*}[!ht]
\footnotesize
\centering
\begin{tabular}{p{8cm}p{8cm}}\\\hline
{\it Input}: hear it ?& {\it RL model choice}: Iter1\\
{\bf Iter1: i 'm sorry i didn 't hear you .}&{\it Iter2}: it 's a little early in the morning for me .\\
{\it Iter5}: that 's the only one who could ever be .&{\it Iter7}: what 's the meaning of this ?\\\hline
{\it Input}: it feels like i must have been asleep for weeks .&{\it RL model choice}: Iter1 \\
{\bf Iter1: i 'm sorry i can 't help you .}&{\it Iter2}: it 's been so long .\\
{\it Iter5}: and then i felt sorry about it .&{\it Iter7}: i 'm sorry i woke you up \\\hline

{\it Input}: we lived in railroad apartments inside red brick . &  {\it RL model choice}: Iter1\\
{\bf Iter1: oh, really ?}&{\it Iter2}: we got a lot of stuff in the trunk .\\
{\it Iter5}: we got to get back to the hotel and get a job .&{\it Iter7}: i lived in a hotel .\\\hline

{\it Input}: you know you ought to find yourself a girl .& {\it RL model choice}: Iter1\\
{\bf Iter1: i 'm not going to do that .}&{\it Iter2}: you know i 'm a real looker .\\
{\it Iter5}: i 'm gonna make you some of your own kind of girl .&{\it Iter7}: you 're in a big house !\\\hline

{\it Input}: the man who does that is not afraid to die .& {\it RL model choice}: Iter1 \\
{\bf Iter1: i am not afraid of him .}&{\it Iter2}: but he 's not afraid of the truth .\\
{\it Iter5}: the man is a man of god who is afraid of god&{\it Iter7}: it is my duty to protect the father .\\\hline
{\it Input}: the cellar he was working on something in the cellar .&{\it RL model choice}: Iter2\\
{\it Iter1}: i was in the cellar .&{\bf Iter2: i 'm going to the cellar .}\\
{\it Iter5}: there 's no way he could have gotten into this room .&{\it Iter7}: we had the whole room .\\\hline
{\it Input}: this little lady seems to be alone in the world .& {\it RL model choice}: Iter2\\
{\it Iter1}: i 'm not going to be able to see her .&{\bf Iter2: i 'm sure she 'll have a good time with the rest of the crew .}\\
{\it Iter5}: you know i can tell when she is a little bit ... ... but she 's not very good&{\it Iter7}: but she 's not coming here .\\\hline
{\it Input}: i didn 't even see you go back there .&  {\it RL model choice}: Iter2\\
{\it Iter1}: i was just going to see you .&{\bf Iter2: i didn 't see the man .}\\
{\it Iter5}: i got to see your father .&{\it Iter7}: i went back to sleep .\\\hline

{\it Input}: i was kinda hoping that you were my dad .& {\it RL model choice}: Iter2\\
{\it Iter1}: i was just wondering if you could help me out .&{\bf Iter2: you 're the only one who 's ever been in my family .}\\
{\it Iter5}: he 's been a long time since he got here .&{\it Iter7}: so what are you doing with my mom ?\\\hline
{\it Input}: excuse me but how did we pay for these suits ?& {\it RL model choice}: Iter2 \\
{\it Iter1}: i don 't know what you are talking about .&{\bf Iter2: i don 't know if i can get a refund on this .}\\
{\it Iter5}: i can 't afford this .&{\it Iter7}: i 'll pay them 10 000 .\\\hline
{\it Input}: maybe his eyes were too big for his stomach. &{\it RL model choice}: Iter2 \\
{\it Iter1}: he was a big boy .&{\bf Iter2: he 's got a big mouth .}\\
{\it Iter5}: i think he was a good kid .&{\it Iter7}: he was just trying to kill her .\\\hline
{\it Input}: like the death zone or the zone of no return . &{\it RL model choice}: Iter2\\
{\it Iter1}: i don 't know what to do .&{\bf Iter2: the zone is the only place where the people are .}\\
{\it Iter5}: the man is a man of the universe .&{\it Iter7}: but it 's a life zone .\\\hline
{\it Input}: home sir .&  {\it RL model choice}: Iter5\\
{\it Iter1}: that 's great !&{\it Iter3}: the rest of the world is here .\\
{\bf Iter5: the rest of the crew are also on their way home  .}&{\it Iter7}: how 's your wife ? .\\\hline
{\it Input}: i 'm sorry to detain you for so long .& {\it RL model choice}: Iter5\\
{\it Iter1}: i 'm sorry i can 't .&{\it Iter2}: i 'm sorry about the delay .\\
{\bf Iter5: i 'm not leaving until tomorrow .}&{\it Iter7}: i 'm glad that 's your name and all .\\\hline
{\it Input}: don 't fuckin talk to me that way man .&{\it RL model choice}: Iter7\\
{\it Iter1}: i 'm not gonna talk to you .&{\it Iter2}: i 'm talkin to you motherfucker .\\
{\it Iter5}: i 'm just sayin that i 'm an asshole and that ain 't the way i talk !&{\bf Iter7: you 're in no position to talk like that .}\\\hline
{\it Input}: can 't you see how they 're exploiting you ?& {\it RL model choice}: Iter7\\
{\it Iter1}: i 'm not a human being .&{\it Iter2}: you 're not a good guy .\\
{\it Iter5}: you 're not an idiot anymore .&{\bf Iter7: well it 's not that simple .}\\\hline
{\it Input}: do you have any idea what caused the explosion ?& {\it RL model choice}: Iter7\\
{\it Iter1}: i have no idea .&{\it Iter2}: i think it 's the same as a gas pipe .\\
{\it Iter5}: i 've never heard of any of these .&{\bf Iter7: the explosion was caused by a meteor disturbance .}\\\hline
\end{tabular}
\caption{Sample of outputs generated by models from different rounds (1,2,5,7) of data distillation, along with the model that the proposed reinforcement learning (RL) algorithm chooses to decode the response. Responses chosen by the RL model are in {\bf Bold}.}
\label{samples}
\end{table*}

\paragraph{Perplexity and diversity}
The size of the training dataset after each round of data distillation and the perplexity of the corresponding trained models on the full development set is shown in the first two columns of Table~\ref{ppl-diverse}. Perplexity increases for models
trained with more data distillation (as expected, since distillation removes
opportunities for the model to learn to produce the most common outputs).

However,
we expect models trained with distillation to complement the model trained on
the entire dataset by better modeling more specific outputs. To quantify the
potential of the pool of generation models to complement each other when used
in different contexts,
we also report {\it oracle perplexity} (``oracle-ppl'') as a function of the
number of iterations {\it K}: for each 
example, we identify the generation model (out of {\it Iter1} through
{\it IterK}) that assigns the highest probability to the true output.
Oracle perplexity is the perplexity computed using these maximal probabilities,
instead of the probabilities assigned by any one model.
This is equivalent to the perplexity of a model with an RL policy network that
chooses perfectly every time. We  expect to find that oracle perplexity
on the development set decreases when adding the models trained in the first few 
rounds of data distillation, after which it levels off. This confirms that
there are benefits to be had from choosing smartly among the different models.

Table~\ref{ppl-diverse} also shows a measure of the diversity of generated responses,
namely, the number of distinct unigrams (``div-1'') and bigrams (``div-2'')
in generated responses as a fraction of the total generated tokens, as described in \cite{li2015diversity}. As can be seen, as the data distillation process proceeds and more generic responses are distilled, the system generates increasingly diverse responses.

\paragraph{Distilled responses}
The highest-frequency responses from different rounds of data distillation are shown in Table~\ref{top-frequent}. 
Top responses are more generic for models trained in earlier iterations.
In iteration 1, the top responses are broadly generic statements of uncertainty (``{\it I don't know}", {\it ``I am not sure"}) or agreement ({\it ``i think you are right"} or {\it ``that's a good idea"}), but the meanings of frequent responses start diverging as the distillation algorithm proceeds. 
The number of the occurrences of the top frequent responses from different iterations also validates this point, with the number gradually decreasing.  

Table \ref{samples} presents sampled outputs from the generation models trained after different rounds of the distillation.
Responses from {\it Iter1} are usually generic but safe, 
mostly {\it i don't know what to do/what you are talking about} and {\it that's a good idea}. As the amount of distilled data increases, the corresponding  
model generates
increasingly concrete responses but has a greater risk of 
  outputting confusing or irrelevant responses.   

\subsection{Choosing the correct model for decoding}
Next, we present results from the proposed reinforcement learning model and analyze how it decides 
which model to pick from the pool.\begin{figure}
\centering
\includegraphics[width=2.5in]{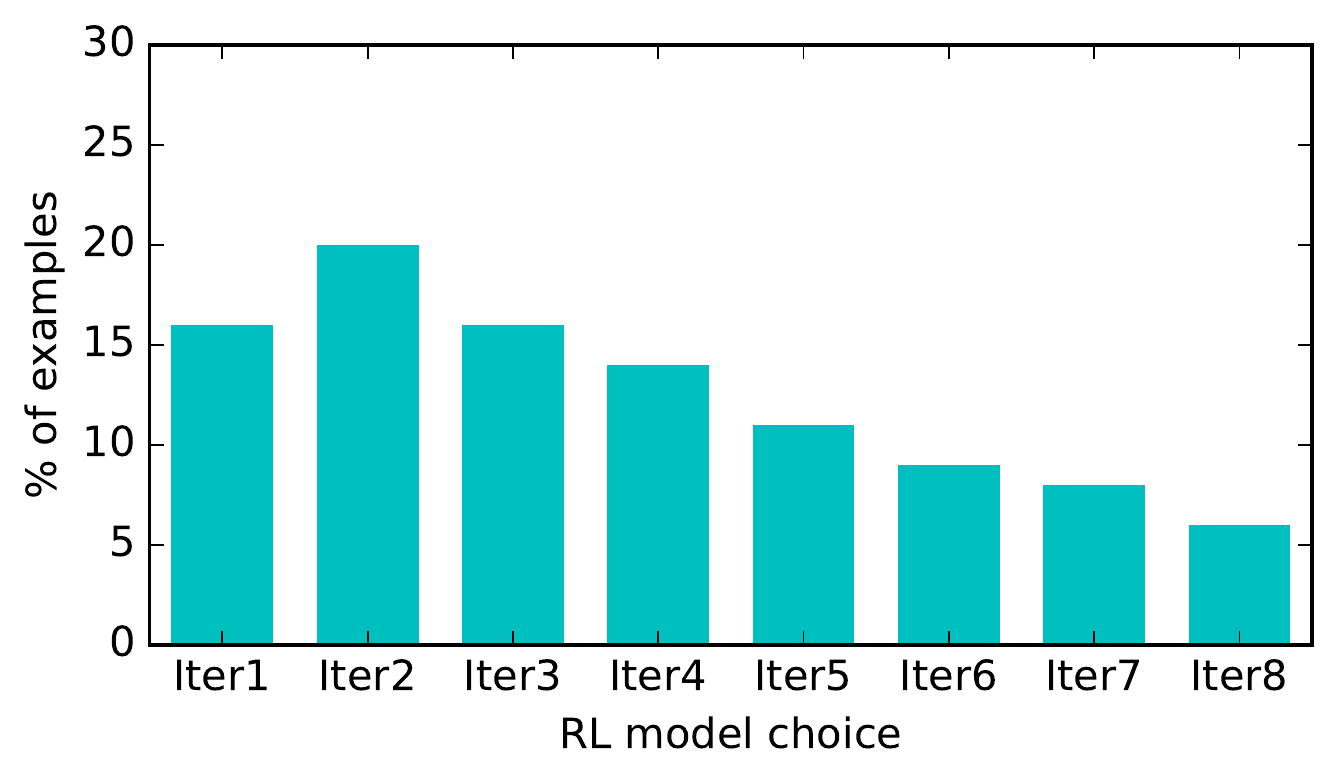}
\caption{Distribution of the iteration used by the RL model
to decode responses (dev set).}
\label{pie}
\end{figure}

The distribution over different models used to decode input messages in the development set is shown in Figure~\ref{pie}. As can be seen, the RL model chooses to decode using the model trained on the entire dataset (i.e., {\it Iter1}) for 16 percent of all inputs. The models trained after 2, 3 and 4 rounds are responsible for decoding responses to approximately half of the inputs. 

\begin{figure}
\centering
\includegraphics[width=2in]{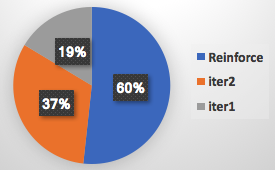}
\caption{The proportions of  outputs from different models ranked first in the human evaluation. Note that the sum is larger than 100 percent due to ties.}
\label{pie-human}
\end{figure}

\begin{table}[!t]
\centering
\small
\begin{tabular}{cccc}
\hline
pairs&win&lose&tie\\\hline
RL vs. Iter2&51&29&20\\
RL vs. Iter1&64&28&8\\
Iter2 vs. Iter1&62&38&-\\\hline
\end{tabular}
\caption{ Pairwise human judgments between the reinforcement learning model ({\it RL}) and the first two distillation models ({\it Iter1} and {\it Iter2}).}
\label{pair}
\end{table}

\paragraph{Human evaluation}
For human evaluation, we follow protocols defined in \newcite{li2016deep}, employing
crowdsourced judges to evaluate a random sample of
200 items. We present labelers with an input message and the
generated outputs from three models, {\it Iter1}, {\it Iter2}, and {\it RL},
and ask them to rank the three outputs by quality. 
Note that the outputs from the {\it RL}
model
 can be the same as those from {\it Iter1} or {\it Iter2} if the {\it RL} model chooses that particular model ({\it Iter1} or {\it Iter2}) for decoding. In these cases, a tie is automatically recorded. 
Figure \ref{pie-human} shows the proportions of the outputs ranked first by the
human labelers.
As can be seen, the reinforcement learning model performs best 60 percent of the time, followed by {\it Iter2}, which wins 37 percent of the time. 
Table \ref{pair} shows pairwise human judgements between the three models extracted from the three-instance ranking. 
It is interesting to see that {\it Iter2} generally outperforms {\it Iter1}, winning on 62 percent of the examples. This is consistent with the fact that 
the {\it RL} model tends to prefer {\it Iter2} more often.

\paragraph{Adversarial evaluation}
Table \ref{adver} reports
 adversarial success and {\it machine-vs-random} accuracy 
   described in \newcite{add}. 
   Adversarial success ({\it AdverSuc}) refers to the percentage of machine-generated responses that are able to fool an trained evaluator model into believe that they are generated by humans; {\it machine-vs-random} accuracy denotes the accuracy of a trained evaluator model (a different evaluator from the one used in adversarial success) at distinguishing between machine-generated responses and human utterances randomly sampled without regard for the input. 
      Superior models should obtain higher values of both adversarial success
      and  {\it machine-vs-random} accuracy. We refer readers to \newcite{add} for more details.
       We observe that the {\it RL} model performs better than always using the model trained on the full dataset ({\it Iter1}) or choosing a distillation model at
       random (as one would expect, since the {\it RL} model is trained to optimize adversarial success). 
       
\begin{table}
\small
\centering
\begin{tabular}{lcc}
\hline
model&{\it AdverSuc}&{\it machine-vs-random}\\\hline
Iter1 (standard)&0.058&0.933\\
random&0.056&0.940\\
RL &0.088&0.944\\\hline
\end{tabular}
\caption{Adversarial success and {\it machine-vs-random} accuracy for {\it Iter1} which always generating response using the model trained on the full set, {\it random} which randomly samples a model for generation, and the proposed model. }
\label{adver}
\end{table}

\paragraph{Analyzing results}
Table \ref{samples} shows  
 example choices made of the {\it RL} model in response to different inputs. When  input messages are vague and hard to reply to, the {\it RL} model usually picks {\it Iter1}, which in turn outputs safe responses like ``{\it that 's great}" or ``{\it i don 't know what you are talking about}''. 
 The {\it RL} model has a tendency to pick models from the latter stages of distillation training if all of the generation models from the different iterations of distillation are able to output meaningful responses, since models from the later stages output produce more diverse and interesting outputs.
We also observe a high correlation between the number of unknown words in the source sentence and the choice to use {\it Iter1}.

\section{Conclusion}
In this paper, we investigate the language specificity issue in dialogue generation. 
We propose 
a data distillation method, which trains a series of generation models that 
exhibit 
 different levels of specificity and uses a reinforcement learning model to choose the model
best suited for decoding depending on the dialogue context. 

The success of the proposed system confirms the importance of data processing in training a successful open-domain dialogue system. We anticipate that strategies resembling
the one we propose can be used more generally for controlling properties of dialogue generation other than specificity, by training several models on different
subsets of a single dataset that differ in the desired property, and choosing among
these to produce outputs that tailor the quality of interest to the situation at
hand.

\bibliographystyle{sl_natbib}
\bibliography{emnlp2016}
\appendix

\end{document}